\pdfoutput=1

\documentclass[11pt]{article}

\usepackage[final]{acl}

\usepackage{times}
\usepackage{latexsym}

\usepackage[T1]{fontenc}

\usepackage[utf8]{inputenc}

\usepackage{microtype}

\usepackage{inconsolata}

\usepackage{graphicx}

\usepackage{bm}
\usepackage{float}
\usepackage{dsfont}
\usepackage{amsmath}
\usepackage{amssymb}
\usepackage{booktabs}
\usepackage{multirow}
\usepackage{enumitem}
\usepackage{algorithm}
\usepackage{algpseudocode}

\newcommand{\sparagraph}[1]{\smallskip \noindent \textbf{#1}}
\DeclareMathOperator*{\argmin}{argmin}
\DeclareMathOperator*{\softmax}{softmax}
\newfloat{algorithm}{t}{lop}

\title{Humanizing Machine-Generated Content: Evading AI-Text Detection through Adversarial Attack}

\author{Ying Zhou$^{1,2}$, Ben He$^{1,2}$\thanks{Corresponding Author}, Le Sun$^{2,3*}$ \\
${}^{1}$University of Chinese Academy of Sciences, Beijing, China \\
${}^{2}$Chinese Information Processing Laboratory ~ ${}^{3}$State Key Laboratory of Computer Science \\
Institute of Software, Chinese Academy of Sciences, Beijing, China \\
{\tt zhouying20@mails.ucas.ac.cn, benhe@ucas.ac.cn, sunle@iscas.ac.cn}
}

\begin{document}
\maketitle
\begin{abstract}
With the development of large language models (LLMs), detecting whether text is generated by a machine becomes increasingly challenging in the face of malicious use cases like the spread of false information, protection of intellectual property, and prevention of academic plagiarism. 
While well-trained text detectors have demonstrated promising performance on unseen test data, recent research suggests that these detectors have vulnerabilities when dealing with adversarial attacks such as paraphrasing.
In this paper, we propose a framework for a broader class of adversarial attacks, designed to perform minor perturbations in machine-generated content to evade detection. 
We consider two attack settings: white-box and black-box, and employ adversarial learning in dynamic scenarios to assess the potential enhancement of the current detection model's robustness against such attacks.
The empirical results reveal that the current detection models can be compromised in as little as 10 seconds, leading to the misclassification of machine-generated text as human-written content.
Furthermore, we explore the prospect of improving the model's robustness over iterative adversarial learning. Although some improvements in model robustness are observed, practical applications still face significant challenges.
These findings shed light on the future development of AI-text detectors, emphasizing the need for more accurate and robust detection methods. 
\end{abstract}

\section{Introduction}

Large language models (LLMs)~\cite{DBLP:journals/corr/abs-2303-08774, DBLP:journals/corr/abs-2305-10403, DBLP:journals/corr/abs-2307-09288} have rapidly emerged as a dominant force within the field of natural language processing (NLP). 
These models acquire extensive internal knowledge through pre-training on large-scale self-supervised data, endowing them the capacity to tackle various tasks, from answering factual questions to generating fluent text, and even performing complex reasoning, which has significantly impacted diverse NLP application domains.
However, these advancements have also raised ethical concerns on their inherent risks~\cite{DBLP:journals/corr/abs-2305-14552, DBLP:journals/corr/abs-2305-04812, DBLP:journals/corr/abs-2304-03738}, including the spread of misinformation, the hallucinations in generated content, and even potential discrimination against specific groups.
The growing recognition of these issues has led to the development of AI-text detection research. Nevertheless, AI-text detector may inherit vulnerabilities from neural network models~\cite{DBLP:journals/corr/SzegedyZSBEGF13}, spurring related research~\cite{DBLP:journals/corr/abs-2303-11156, DBLP:journals/corr/abs-2303-13408} aimed at conducting paraphrasing attacks on AI detectors to mislead their predictions. 
We believe that the exploration of potential adversarial attacks on text detectors is of paramount importance, as it allows for the identification of vulnerabilities in AI detectors before their deployment in real-world applications, such as student essay plagiarism detection, while also facilitating the development of appropriate countermeasures.

Current detection methods are typically categorized into three groups: those relying on statistical measures~\cite{DBLP:conf/icml/Mitchell0KMF23} like entropy, perplexity, and log-likelihood; those training neural classifiers~\cite{DBLP:journals/corr/abs-2301-07597} from supervised data with human/AI-generated labels; and those utilizing watermarking~\cite{DBLP:conf/icml/KirchenbauerGWK23} to inject imperceptible pattern to the AI-generated text.
Unfortunately, limited research has explored the adversarial perturbations targeting AI-text detectors. Notably, \citet{DBLP:journals/corr/abs-2303-11156, DBLP:journals/corr/abs-2303-13408} explored the use of paraphraser to rewrite machine-generated content for adversarial attacks. Simultaneously, \citet{DBLP:journals/corr/abs-2305-19713} utilized LLMs to generate adversarial word candidates, and created adversarial results using a search-based method.
While these prior studies have revealed the vulnerabilities of AI detectors to adversarial perturbations, the influence of adversarial attacks on the detector in real-world dynamic scenarios remains largely unexplored.

\begin{figure}[t]
\centering
\includegraphics[scale=0.24]{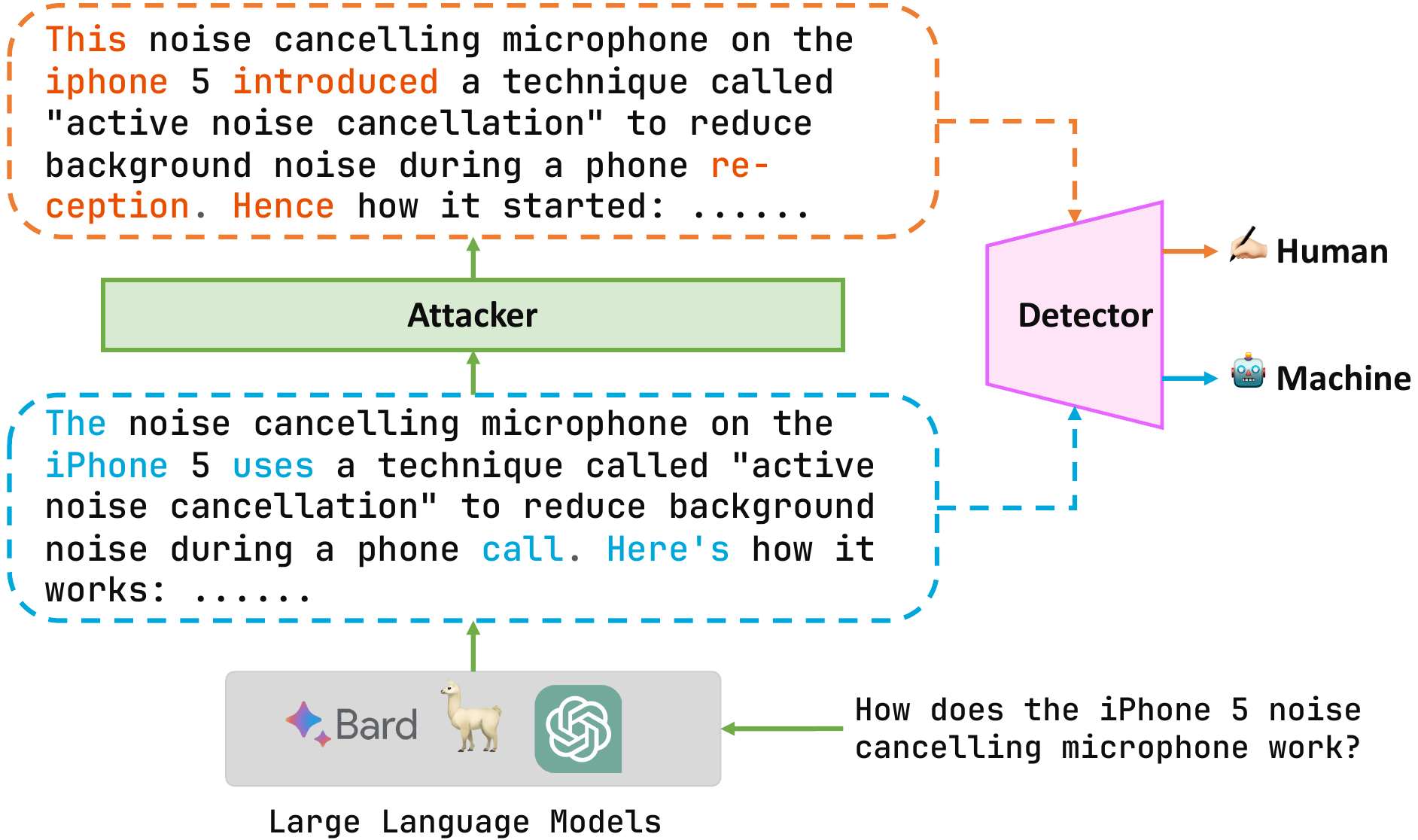}
\caption{Overview of the Adversarial Detection Attack on AI-Text (ADAT) task.} 
\label{fig:adat} 
\end{figure}

In this paper, we propose a broader task: Adversarial Detection Attack on AI-Text (ADAT).
The objective of ADAT is to perturb AI-generated text in a semantically preserving manner, thereby influencing the detector's predictions and enabling machine-generated text to evade detection. 
Figure~\ref{fig:adat} outlines the general process of ADAT, along with an example of attacks\footnote{Sample in figure~\ref{fig:adat} is from real AI-text attacks, available for testing at \url{https://huggingface.co/Hello-SimpleAI/chatgpt-detector-roberta}}.
We consider both black-box attack setting, where the attacker can access only the target detector predictions without any internal information of the detection model, and white-box attack setting. To bridge the gap between these settings, we employ an intermediate model to unify the attack methods.
Additionally, to address the challenge of constructing a more robust detection model, we introduce an adversarial learning approach in a dynamic scenario for the ADAT task, whereas the detector iteratively updates its parameters using adversarial samples. Our experiments validate the enhanced robustness achieved through this approach and highlight potential challenges.
Building on the above considerations, we introduce a novel and comprehensive framework called the Humanizing Machine-Generated Content (HMGC) for ADAT tasks, which is designed to facilitate the interaction process between the attacker and the detector.

Our main contributions could be summarized as follows:
\begin{itemize}[leftmargin=*, itemsep=-0.3em, topsep=0.2em]
\item To the best of our knowledge, ADAT is the first rigorously defined task in the field of adversarial attacks on AI-text detection. It encompasses both white-box and black-box attack settings, serving as a foundational reference for future research in this domain.
\item We introduce the HMGC framework, which offers a general attack paradigm for ADAT tasks. Extensive experiments reveal the efficacy of the HMGC framework. In particular, we provide empirical evidence highlighting the significance of perplexity in AI-text detection.
\item Our experimental results demonstrate that the proposed adversarial learning approach in dynamic scenarios effectively enhances the robustness of detection models, suggesting the potential to train a universal AI-text detector through dynamic adversarial learning. The data and related resources are available online\footnote{\url{https://github.com/zhouying20/HMGC}}.
\end{itemize}

\section{Related Works}

\sparagraph{Large Language Model.}
The powerful capabilities of large language models, exemplified by GPT~\cite{DBLP:journals/corr/abs-2303-08774}, PaLM~\cite{DBLP:journals/corr/abs-2305-10403}, and LLaMA~\cite{DBLP:journals/corr/abs-2307-09288} have revolutionized the application landscape within the field of natural language processing. 
These models can generate coherent and fluent text enriched with external knowledge, effectively tackling complex issues across various domains, from physics~\cite{DBLP:journals/corr/abs-2303-01067} and medicine~\cite{DBLP:journals/jms/DiGiorgioE23} to mathematics~\cite{DBLP:conf/icitl/LiLCSH23} and  linguistics~\cite{DBLP:journals/corr/abs-2304-01852}.
Nonetheless, current language models still grapple with issues like hallucination~\cite{DBLP:journals/corr/abs-2305-14552}, the inadvertent spread of misinformation~\cite{DBLP:journals/corr/abs-2305-04812}, and the potential for value discrimination~\cite{DBLP:journals/corr/abs-2304-03738} in practical applications. 
Consequently, the regulation of large language models to mitigate the risk of significant social problems has become increasingly vital.
We focus on studying how AI-generated text can circumvent existing detection mechanisms, aiming to provide valuable insights and perspectives for the development of robust detection models.

\sparagraph{AI-text Detection.}
Current text detection methods typically fall into three categories:
1) Statistical methods~\cite{DBLP:conf/acl/GehrmannSR19, DBLP:conf/ecai/LavergneUY08, DBLP:journals/corr/abs-1908-09203, DBLP:conf/icml/Mitchell0KMF23, DBLP:journals/corr/abs-2306-05540} employ statistical tools to make zero-shot distinctions between human and machine-generated text using measures such as information entropy, perplexity, and n-gram frequency.
Recent work in this category includes DetectGPT~\cite{DBLP:conf/icml/Mitchell0KMF23}, which observed that text generated by language models often resides in the negative curvature region of the log probability function, and it proposed the defining curvature-based criteria to distinguish AI-text.
2) Classifier-based methods involve training text detection models based on supervised data~\cite{DBLP:conf/emnlp/UchenduLSL20, DBLP:journals/corr/abs-2305-16617, DBLP:journals/corr/abs-2305-09859}, whereas recent research~\cite{DBLP:journals/corr/abs-2301-07597, DBLP:journals/corr/abs-2306-05524} often utilizes RoBERTa to train a binary classifier for text detection. 
However, this method requires a substantial amount of training data and faces challenges related to weak generalization~\cite{DBLP:journals/corr/abs-1906-03351} and limited robustness against attacks~\cite{DBLP:journals/corr/abs-2303-14822, DBLP:conf/acl/QiLCZ0WS20}.
3) Watermarking methods are emerging with the rise of decoder-only large language models, which imprint specific patterns on generated text.
For instance, \citet{DBLP:conf/icml/KirchenbauerGWK23} propose randomly partitioning the vocabulary into a green list and a red list during text generation, with the division based on the hash of previously generated tokens. Meanwhile, \citet{DBLP:journals/corr/abs-2307-16230} introduces a watermarking method akin to the RSA asymmetric cryptography. 
Challenges for this category include that the generated text may lack smoothness, and the watermark pattern is susceptible to leakage~\cite{DBLP:journals/corr/abs-2307-16230}.

\sparagraph{Adversarial Attack.}
A few recent works have made attempts to attack text detection models. 
For example, \citet{DBLP:journals/corr/abs-2303-11156, DBLP:journals/corr/abs-2303-13408} proposed to use paraphrasers to rewrite the generated text of LLMs, successfully evading detection by the 3 categories of models mentioned above. 
Notably, our work shares common views on the usage of adversarial learning with~\citet{DBLP:journals/corr/abs-2307-03838}, but differs in that~\citet{DBLP:journals/corr/abs-2307-03838} introduced a paraphraser to enhance the robustness of detector, whereas our work explores whether the detector can continue to learn from multiple rounds of attacks in dynamic scenarios to resist adversarial attacks.
\citet{DBLP:journals/corr/abs-2305-19713} demonstrated the utility of word substitution attacks against AI-text detectors. When compared to our approach, it's worth noting that their method relies on LLMs to generate candidate words, which might encounter efficiency challenges when computing resources are limited.
Building upon these observations, our paper presents a comprehensive detector adversarial attack framework called HMGC. The key distinction between our work and these studies lies in our introduction of the ADAT task, providing a formal paradigm for future research. 
Moreover, we also emphasize our work as universal adversarial perturbations, which can be applied to any input for any target detector model.

\section{Preliminary}
In this section, we introduce the key definitions of the Adversarial Detection Attack on AI-Text (ADAT) task.

\subsection{Problem Statement} \label{sec:problem_statement}
In the detection of machine-generated content, when presented with a set of user requirements $\mathcal{U}$ and their corresponding response articles $\mathcal{T} = \mathcal{T}_{human} \cup \mathcal{T}_{machine}$, the objective of the detection model $\mathcal{D}$ is to assign a score $\mathcal{D}(u, t)$ to each article to help users discern whether the article is generated by machines, specifically for large language models (LLMs) like ChatGPT or Bard. 
For instance, given a threshold of 0.5, the detector generates detection probabilities for all articles, resulting in an array $\mathcal{P}=[D_1, D_2, ..., D_n]$, in which articles with a detection probability exceeding 0.5 are identified as machine-generated content, defined as $\overline{\mathcal{T}}_{machine} = \{t\ |\ D(u_t, t) > 0.5\}$.

\sparagraph{Adversarial Attack on Detection.}
We propose the task of \underbar{A}dversarial \underbar{D}etection \underbar{A}ttack on AI-\underbar{T}ext (ADAT). The objective is to introduce subtle modifications to machine-generated articles, aiming to fool the detector into classifying them as human-authored. 
Formally defined, given user requirements $u_t$ and a machine-generated article $t$, the attack's objective is to construct an effective adversarial sample $t_{adv}$ based on $t$, ensuring its detection probability falls below the classification threshold. 
Concretely, we define a successful attack as: $D(u_t, t_{adv}) < threshold$ when $D(u_t, t) \geq threshold$, with the condition that $Dis(t, t_{adv}) \leq \epsilon$. 
Here, $D(u_t, t)$ and $D(u_t, t_{adv})$ are the detection probabilities of the text before and after the attack, respectively. $Dis$ is the similarity distance evaluation function, and $\epsilon$ is a small value ensuring minimal deviation of the text distribution in the adversarial sample from the original text.

\sparagraph{White-box and Black-box Attack.}
In terms of attack methodologies against the detector, we establish two realistic settings:
In the \textbf{white-box attack} scenario, the attacker accesses comprehensive information about the victim detector, including parameters, gradients, training data, and more.
Contrastingly, in the \textbf{black-box attack} scenario, which aligns more closely with practical applications, the attacker can only access the output results of the victim model. This means only $\mathcal{D}(u, t_{adv})$ or even the binary predictions are provided. Typically, this is further categorized into score-based black-box attacking and decision-based black-box attacking~\cite{DBLP:journals/tois/WuZGRFC23}.
In this work, we both consider \textit{white-box attacks} and \textit{decision-based black-box attacks}.

\begin{table}[t]
\centering         
\resizebox{0.99\linewidth}{!} {%
\begin{tabular}{l|*{2}{c}}
\toprule
& {\bfseries CheckGPT} & {\bfseries HC3}\\
\midrule
Training size & 720,000* & 58,508 \\
Testing size & 90,000* & 25,049 \\
Avg \#words  & 136.68 & 145.89 \\
Domains & News, Essay, Research & QA \\
\bottomrule
\end{tabular}
}
\caption{Data statistics, where * denotes the data are randomly split with seed 42, and \#words denotes the number of words in one sample.}
\label{tab:data}
\end{table}

\sparagraph{Dynamic Adversarial Attack}
Another crucial aspect of the detector is its ability to undergo continuous updates using augmentation data from users or other models. 
As mentioned above, \citet{DBLP:journals/corr/abs-2307-03838} proposed leveraging a paraphraser as a data generation method to fortify the detector against rewriting attacks during training, resulting in sufficiently robust and transferable detection results. However, this approach remains confined to the model training phase and doesn't explore the robustness and adaptability of the detector in a more dynamic scenario. 
In this work, we introduce the concept of \textbf{dynamic attacks} to iteratively optimize the interplay between attacker and detector across multiple rounds. Detailed processes are shown in Section~\ref{sec:method}.

\subsection{Benchmarking}
To validate the performance of adversarial attacks in different scenarios, we selected two datasets:
1) \textit{CheckGPT}~\cite{DBLP:journals/corr/abs-2306-05524}:
Due to the unavailability of the training data division or other details of its detector except for the detector model checkpoint and the full dataset, we consider this scenario as a black box attack, and we partition the entire dataset randomly into 80\% training set, 10\% validation set, and 10\% test set. 
A surrogate model is trained to act as a proxy for the attacker, and subsequently, we assess the effect of the black box attack on the released original model.
2) \textit{HC3}~\cite{DBLP:journals/corr/abs-2301-07597}:
The data partitioning and model parameters are publicly available, making it suitable for a white-box attack. 
We utilize the public test set as the attack samples and employ the released classifier as the victim model to evaluate the effectiveness of the white-box attack. 
Further details about both datasets are presented in Table~\ref{tab:data}.

\section{Methodology} \label{sec:method}
In this section, we start by providing an overview and mathematical definitions of our proposed attack methodologies. 
Following this, we outline the detailed process of establishing a unified framework that bridges the gap between black-box and white-box attack scenarios in Section \ref{sec:method-surrogate} through a surrogate victim model. 
Furthermore, we delve into the core of our adversarial attack method in Sections \ref{sec:method-constraints}, \ref{sec:method-wir}, and \ref{sec:method-mlm} by elucidating the constraints, word importance, and word replacement strategy. 
Finally, in Section \ref{sec:method-dynamic}, we introduce our innovative evaluation paradigm focused on dynamic adversarial attacks.

\sparagraph{Algorithm Overview.} 
The HMGC framework we introduce can be conceptualized as an ongoing interaction between the attacker and the detector.
When presented with machine-generated text, the attacker iteratively modifies the text in an attempt to fool the detector. This process continues until the detector finally classifies the adversarial text as human-generated.
We illustrate the general process in pseudo Algorithm~\ref{alg:overview}.
More specifically, the attacker in our HMGC framework comprises four key concepts: the surrogate detection model $\mathcal{D_{\theta}}$, the word importance ranker $\mathcal{R}$, the encoder-based word swapper $\mathcal{M}$, and a set of constraints $\mathbb{C}=\{c_1, c_2, ..., c_k\}$. 
The final objective of the ADAT task could be formulated as:
\begin{equation}
\begin{gathered}
t_{adv} = \argmin_{t_{adv}} \mathcal{D_{\theta}}(t_{adv}), \\
s.t.~~~t_{adv}\in \{t_{tgt}\} \cup \mathcal{M}(t_{tgt}, R(w_i)), \\
\sum_{c_i \in \mathbb{C}} \mathds{1}\left(c_i(t_{adv})\right) = |\mathbb{C}|.
\end{gathered}
\end{equation}

\begin{algorithm}[htb!]
\caption{HMGC}
\label{alg:overview}

\textbf{Input:} the original detection model $\mathcal{D}_{ori}$, pre-collected training dataset $\mathbf{T}_C$, a target text $t$, and an encoder model $\mathcal{M}_{mlm}$, a set of attack constraints $\mathbb{C}$ \\
\textbf{Parameter}: $\tau$ threshold for human-written detection, $k$ maximum words can be replaced in one attack \\
\textbf{Output:} adversarial text $t_{adv}$ 

\begin{algorithmic}[1] 
\State Initialize $t_{curr}$ to represent the current text to be attacked: $t_{curr} \leftarrow t$

\State {\bfseries procedure 1.} Train Surrogate Model
\For {$t_c$ in $\mathbf{T}_C$}
    \State $P_{D} \leftarrow $ predict all $t_c$ using $\mathcal{D}_{ori}$
\EndFor
\State Train surrogate model $\mathcal{D}_{\theta}$ on $\mathbf{T}_C$ using $P_{D}$ as the target label in terms of Eq.\ref{eq:surrogate}

\State {\bfseries procedure 2.} Pre-attack Assessment
\State Predict whether the current sample is human-written: $p_{h} \leftarrow \mathcal{D}_{\theta}(t_{curr})$

\State {\bfseries procedure 3.} Word Importance Ranking
\State $W_{curr} = \{w_1, w_2, ..., w_n\} \leftarrow$ Split current sample to words
\State Calculate word importance $I_{w_i}$ based on gradient and perplexity for each word with Eq.\ref{eq:dualir}
\State $W[:k] \leftarrow$ Sort $W_{curr}$ based on importance

\State {\bfseries procedure 4.} Mask Word Substitution
\For{$w_i$ in $W[:k]$}
\State Replace $w_i$ in $t_{curr}$ with MASK token
\State Obtain $M$ candidates $\{p_m\}_{m=1}^M$ with Eq.\ref{eq:substitution}
\State $p_m^* \leftarrow argmax_{m=1}^M(\mathcal{D}_{\theta}(t_{curr}+p_m))$
\State $t_{opt} \leftarrow$ Fill in $p_m^*$ for $t_{curr}$
\If{$\mathcal{D}_{\theta}(t_{opt}) < \mathcal{D}_{\theta}(t_{curr})$}
    \State $t_{curr} \leftarrow t_{opt}$
\EndIf
\State Post-checking for $t_{curr}$ with Eq.\ref{eq:constraint} on $\mathbb{C}$
\EndFor

\State {\bfseries procedure 5.} Post-attack Checking
\If {$\mathcal{D}_{\theta}(t_{curr}) < \tau$ \textbf{or} reach attack limits}
    \State \textbf{return} $t_{adv} = t_{curr}$ \Comment{Algorithm terminates}
\Else
    \State \textbf{goto} procedure 3. \Comment{Repeat the process}
\EndIf

\end{algorithmic}
\end{algorithm}

\subsection{Surrogate Victim Model} \label{sec:method-surrogate}
Under the black-box attack setting, obtaining the internal information of the detection model directly is not feasible. To compute the importance of words, it is necessary to train a surrogate model that emulates the behavior of the original detection model and can provide gradients for adversarial attacks as a proxy.
Following~\cite{DBLP:journals/corr/abs-2301-07597}, we train the surrogate model based on RoBERTa~\cite{DBLP:journals/corr/abs-1907-11692} with the binary classification task. The training supervision signal is distilled directly from the predictions of the original model.
Formally, leveraging a pre-collected training dataset $\mathbf{T}$, we employ the original detection model $\mathcal{D}_{ori}$ to predict each sample, obtaining a set of prediction results, $P_D$. 
Subsequently, we initialize the surrogate detection model $\mathcal{D}_{\theta}$ using the original RoBERTa for training, whereas the training objective is as follows:
\begin{equation} \label{eq:surrogate}
\mathcal{L}_{\theta} = -\big(p_i log(\hat{y}_i)\big)+(1-p_i)\log(1-\hat{y}_i)),
\end{equation}
where $p_i \in P_D$, and $\hat{y}_i = \mathcal{D}_{\theta}(t_i), t_i \in \mathbf{T}$.

\subsection{Word Importance Ranking} \label{sec:method-wir}
In our preliminary experiments, we observed that the detector exhibits greater sensitivity to individual words within the text, particularly those that occurred in user requirements. 
So perturbing important words within the text tends to be more effective in carrying out adversarial attacks. To address this, we have proposed a dual-aspect word importance ranking algorithm that combines model gradients and perplexity derived from large language models.
Firstly, it is intuitive that attacks are more effective on tokens with higher gradients on the victim model, whereas higher gradients indicate greater impacts on the final result. 
Consequently, we consider the gradient norm value corresponding to the \textit{i}-th token as the first aspect of word importance:
\begin{equation}
I_{w_i}^g = \Bigg\| \frac{\partial\mathcal{L}_{\theta}}{\partial\mathbf{e}_{w_i}} \Bigg\|_1,
\end{equation}
where $\mathcal{L}_{\theta}$ is the loss of the objective function; $\mathbf{e}_{w_i}$ is the embedding vector for the \textit{i}-th token in the surrogate model.

Furthermore, existing research~\cite{DBLP:journals/corr/abs-2301-07597, DBLP:journals/patterns/LiangYMWZ23} on AI-text detection has emphasized the importance of language perplexity as a key indicator for distinguishing between human and machine-generated text. Typically, machine-generated content exhibits lower perplexity. 
To enhance the effectiveness of our attacks, we introduce additional constraints aimed at increasing the perplexity of the adversarial results, whereas we propose the use of LLM perplexity as a measure for our word importance ranking. 
More specifically, for each input token, we calculate the perplexity importance as the difference in language perplexity before and after the \textit{i}-th token is removed:
\begin{equation}
I_{w_i}^p = ppl(t_{\backslash w_i})-ppl(t),
\end{equation}
where $t_{\backslash w_i}$ represents the text after removing the \textit{i}-th token.

Subsequently, by introducing $\alpha$ as the weighting factor, we obtain the final word importance score:
\begin{equation} \label{eq:dualir}
I_{w_i} = (1 - \alpha) I_{w_i}^g + \alpha I_{w_i}^p.
\end{equation}

\subsection{Mask Word Replacements} \label{sec:method-mlm}
Here, for an adversarial attack, we sequentially obtain synonymous candidates for each word based on its word importance score in descending order and replace them back into the original text.
For instance, for the i-th important token in the text $t$, we: 1) Replace the token with [MASK]. 2) Utilize the encoder-based model $\mathcal{M}_{mlm}$ to predict the logits for the masked position and perform softmax. 3) Select the top $k$ words with the highest scores as candidates. 4) Replace the candidate words back into the original text one by one, and get the final result following a greedy search strategy.
It's worth noting that the source for synonym generation can be any suitable algorithm, such as word embedding spaces~\cite{DBLP:conf/naacl/MrksicSTGRSVWY16} or querying WordNet\footnote{\url{https://wordnet.princeton.edu/}}.
However, our early experiments have shown that using an encoder-based model is the most effective way for a random replacement.
In summary, we generate candidate synonyms as:
\begin{equation} \label{eq:substitution}
p_m^i = \softmax_{m=1,2,...,k}\left(\mathcal{M}_{mlm}(t_{\backslash w_i}+[MASK]_i)\right).
\end{equation}

\subsection{Attack Quality Constraints} \label{sec:method-constraints}
Following the word replacement process, it can be challenging to ensure that the semantics of the original text remain relatively unchanged. 
For instance, a sentence like ``I like that guy'' might be transformed into ``I hate that guy'' after perturbation, resulting in a complete reversal of sentiment. 
To maintain both syntactic correctness and semantic consistency, we introduce three additional constraints to control the extent of deviation: 
1) POS Constraint enforces that the candidate word must align with the part of speech of the word it's replacing, e.g., adjectives cannot be used to substitute nouns. 
2) Maximum Perturbed Ratio Constraint limits the proportion of replacement words in the original text within a certain threshold. 
3) USE Constraint utilizes the Universal Sentence Encoder (USE)~\cite{DBLP:conf/emnlp/CerYKHLJCGYTSK18} as a sentence similarity scorer to measure the distance between the context window of the replacement word and the original text to address the possible semantic shift problem. If the difference is too substantial, the attack is abandoned.
In formal terms, for each constraint $c_i$ and the current adversarial text $t_{adv}$:
\begin{equation} \label{eq:constraint}
c_i(t_{adv}) = 
\begin{cases}
true &\text{if $t_{adv}$ satisfies $c_i$}\\
false &\text{else}
\end{cases}
\end{equation}

\subsection{Dynamic Detector Finetuning} \label{sec:method-dynamic}
As mentioned in Section~\ref{sec:problem_statement}, in the dynamic attack setting, following each round of attacks that gather a substantial collection of adversarial samples, we proceed to continue training the surrogate model in terms of Eq.~\ref{eq:surrogate}. 
This process is designed to strengthen the detector's defense capabilities against one specific form of attack, thereby enabling us to simulate a real-world application scenario where the detector accumulates user queries and continually enhances its capabilities.

\section{Experiments}
In this section, we first introduce our experimental setup. 
Next, we compare the performance between HMGC and the baselines in both black-box and white-box attack settings. 
We then move to the dynamic attack setting, conducting 10 rounds of attack-then-detect iterations to assess the impact of adversarial learning on attack efficacy. 
Finally, we conduct ablation experiments to analyze the significance of different modules within HMGC.

\subsection{Experimental Setup}

\subsubsection{Evaluation Metrics}
\sparagraph{Attack performance measures.} In line with previous research~\cite{DBLP:conf/icml/Mitchell0KMF23}, we employ the AUC-ROC and the confusion matrix to evaluate the attack performance:
1) {\bfseries AUC-ROC} is a performance measure that assesses the area under the receiver operating characteristic curve,
whereas a higher AUC-ROC score indicates better detection performance.
2) {\bfseries Confusion matrix} provides a detailed breakdown of the model's performance, with 'positive' denoting human-written content. We report the following three metrics: 
Positive predictive value (PPV) $\frac{TP}{TP+FP}$, i.e., the proportion of human-written cases among all predicted cases classified as human-written.
True negative rate (TNR) $\frac{TN}{TN+FP}$, i.e., the accuracy in classifying machine-generated text. 
We also denote the decrease of TNR as $\Delta$Acc, which quantifies the reduction in the accuracy of machine-generated sample detection after the attack. It is calculated as $\frac{\text{TNR before Attack} - \text{TNR after Attack}}{\text{TNR before Attack}}$.

\sparagraph{Text quality measures.} Here, we use the following metrics to evaluate the text quality after the adversarial attack. 
\textbf{Flesch reading ease}: Higher Flesch scores indicate that the material is easier to read. 
To assess the impact of text's readability, we use the difference ratio of the Flesch score, denoted as $\Delta \text{Flesch}\%$.
\textbf{Perplexity from LLMs}: It measures the level of uncertainty of a given document. We calculate the change in perplexity before and after the attack with Pythia-3B~\cite{DBLP:conf/icml/BidermanSABOHKP23} to measure the overall quality of an adversarial text, denoted as $\Delta \text{ppl}$.

\subsubsection{Baselines}
Referring to recent research~\cite{DBLP:journals/corr/abs-2303-11156, DBLP:journals/corr/abs-2303-13408, DBLP:journals/corr/abs-2305-19713}, we investigate three primary categories of baseline algorithms for the ADAT task: 
{\bfseries Word-level perturbation} treat tokens in the input as the smallest attack units. It disrupts the detection model by substituting specific words in the original text, typically with words that have similar meanings. We consider {\itshape WordNet} and {\itshape BERT MLM predictions} as the sources for synonyms. 
{\bfseries Sentence-level perturbation} commonly employs a seq-to-seq model to rephrase or rewrite sentences from the original text, thereby perturbing the distribution of the original content. In our study, we examined two strategies serving as baselines: introducing {\itshape irrelevant sentences} and utilizing {\itshape BART} to replace random sentences from the original text.
{\bfseries Full-text rewriting perturbation} involves using a rewriter to directly substitute the original text, effectively evading detection. We considered three methods: {\itshape back translation} which translates the original English text into German and then translates it back to English, and crafting the prompt to instruct a {\itshape LLaMA-2} to rewrite the article, aiming to maximize the divergence from the original text. Moreover, we also employ the SoTA paraphraser {\itshape DIPPER}~\cite{DBLP:journals/corr/abs-2303-13408} with the lex=40, order=40, the most effective setting in their paper.

\subsubsection{Model Ablations}
As discussed in Section \ref{sec:method}, four variations of the HMGC model are implemented by modifying the constraint module and the word importance calculation method:
1) HMGC.$_{-POS}$ does not enforce the replacement and original words to belong to the same part of speech.
2) HMGC.$_{-USE}$ eliminates the constraints related to semantic space consistency using USE~\cite{DBLP:conf/emnlp/CerYKHLJCGYTSK18}.
3) HMGC.$_{-MPR}$, where there is no restriction on the proportion of words that can be replaced.
4) HMGC.$_{-PPL}$ removes the constraint of semantic perplexity in the word importance method, relying solely on the gradient information of the victim model.

\begin{table*}[t]
\centering         
\resizebox{0.99\linewidth}{!} {%
\begin{tabular}{l|*{4}{c}|*{4}{c}|*{1}{c}}
\toprule
\multirow{2}{*}{\bfseries Model} & \multicolumn{4}{c}{\textbf{White-box Attack on HC3}} & \multicolumn{4}{c}{\textbf{Black-box Attack on CheckGPT}} & {\bfseries Duration}\\
& {\bfseries AUC}~$\downarrow$ & {\bfseries PPV}~$\downarrow$ & {\bfseries TNR}~$\downarrow$ & {\bfseries $\bm{\Delta}$Acc}~$\uparrow$ & {\bfseries AUC}~$\downarrow$ & {\bfseries PPV}~$\downarrow$ & {\bfseries TNR}~$\downarrow$ & {\bfseries $\bm{\Delta}$Acc}~$\uparrow$ & {\bfseries Sec/Sample}~$\downarrow$ \\
\midrule
{WordNet Syn} & 98.36 & 98.73 & 97.30 & 2.55 & 91.39 & 85.56 & 83.07 & 16.81 & $\approx0$ \\
{MLM Syn} & 97.79 & 98.19 & 96.15 & 3.70 & 87.68 & 80.46 & 75.65 & 24.24 & 0.1 \\
{Irr Sent} & 98.66 & 99.00 & 97.89 & 1.96 & 95.88 & 92.66 & 92.06 & 7.81 & $\approx0$ \\
{MLM Sent} & 95.80 & 96.40 & 92.17 & 7.69 & 94.26 & 89.97 & 88.83 & 11.05 & 5.27 \\
{Back Trans} & 99.20 & 99.51 & 98.97 & 0.87 & 94.89 & 90.99 & 90.07 & 9.80 & 3.26 \\
{LLaMA-2-7B} & 95.94 & 96.52 & 92.45 & 7.41 & 96.13 & 93.09 & 92.56 & 7.31 & 9.51 \\
{LLaMA-2-13B} & 97.97 & 98.37 & 96.52 & 3.33 & 96.23 & 93.26 & 92.76 & 7.11 & 10.61 \\
{DIPPER} & 98.62 & 98.78 & 97.82 & 2.02 & 88.77 & 81.90 & 77.84 & 22.05 & 14.43 \\
\midrule
{\bfseries HMGC} & {\bfseries 51.06} & {\bfseries 68.29} & {\bfseries 2.70} & {\bfseries 97.29} & {\bfseries 76.64} & {\bfseries 68.35} & {\bfseries 53.57} & {\bfseries 46.35} &  9.25  \\
\bottomrule
\end{tabular}
}
\caption{Attack performance of white-box and black-box setting on HC3 and CheckGPT.}
\label{tab:main}
\end{table*}

\subsubsection{Implementation Details}
For the black-box attack using CheckGPT datasets, we adopt RoBERTa as the surrogate model, where we distilled the original detection performance over two epochs on the 720k training data. 
In detail, the maximum sequence length is set to 512, and the learning rate is set to 5e-6.
As for both white-box and black-box attacks, we select 10k samples from their test set for attacking.
In the dynamic attack, we randomly divide the 90k test data into 10 equal parts. After each attack, 80\% of the attack results are incorporated as new training data, while 20\% of the attack results (equivalent to 1.8k) are utilized for evaluating the model.
For the attacker, the perplexity weighting factor $\alpha$ is set at 0.2, the window size for the USE in the fluency constraint is 50, the minimum tolerance threshold $\gamma$ is 0.75, and the maximum proportion of replaceable words does not exceed 40\% of the original text. 
All experiments were conducted on a machine equipped with six 3090 GPUs.

\begin{table}[t]
\centering         
\resizebox{0.99\linewidth}{!} {%
\begin{tabular}{c|*{4}{c}|*{1}{c}}
\toprule
\multirow{2}{*}{\bfseries Round} & \multicolumn{4}{c}{\textbf{Automatic Metrics}} & {\bfseries Duration}\\
& {\bfseries AUC} & {\bfseries PPV} & {\bfseries TNR} & {\bfseries $\bm{\Delta}$Acc} & {\bfseries Sec} \\
\midrule
{1} & 49.05 & 49.18 & 0.44 & 99.56 & 9.25 \\
{2} & 46.64 & 49.06 & 1.69 & 98.30 & 15.11 \\
{3} & 48.47 & 45.85 & 4.79 & 95.21 & 20.99 \\
{4} & 62.87 & 56.61 & 32.61 & 67.35 & 26.40 \\
{5} & 75.89 & 68.61 & 58.91 & 40.83 & 30.40 \\
{6} & 83.16 & 80.62 & 78.27 & 21.29 & 32.30 \\
{7} & 88.58 & 87.12 & 87.09 & 12.53 & 33.10 \\
{8} & 90.00 & 90.17 & 90.23 & 9.47 & 33.40 \\
{9} & 90.80 & 87.30 & 87.25 & 12.19 & 33.17 \\
{10} & 87.25 & 85.96 & 85.06 & 14.85 & 32.00 \\
\bottomrule
\end{tabular}
}
\caption{
Attack performance in a dynamic environment on CheckGPT. The first-round model represents the original surrogate model, evaluated on test data generated by its adversarial attacks. The second-round model signifies the model that has undergone an adversarial learning process and is evaluated on the test data generated by its adversarial attack. This process continues iteratively for subsequent rounds.
}
\label{tab:dynamic}
\end{table}

\subsection{Experimental Results}

\sparagraph{Detectors are vulnerable to adversarial attack.}  
Table~\ref{tab:main} shows the results of both the baseline models and our proposed HMGC in white-box and black-box attack settings for the two datasets.
Notably, white-box attacks naturally provide precise insights into the internal information of the detection model, making them less robust against disturbance. 
The AUC, for instance, demonstrates a significant drop in detection performance, plummeting from 99.63\% pre-attack to a mere 51.06\%, which is akin to that of a random binary classifier. It should be noted that the HC3 dataset comprises more human-generated articles (67.82\%), leading to a PPV much higher than 50\%. 
Meanwhile, for a more intuitive measure $\Delta$Acc, after the attack, the model's misclassification rate for machine-generated content surged by 97.29\%.
Furthermore, in the more challenging black-box attack setting, our proposed HMGC demonstrated considerable effectiveness.
It successfully perturbed approximately 46\% of machine-generated articles, marking a 22\% improvement compared to the optimal baseline model.
In general, regardless of whether in white-box or black-box attack settings, our proposed HMGC method consistently outperforms all the baseline models.

\sparagraph{Training method for the detector significantly influences its robustness.}
Analyzing the baseline performance from Table~\ref{tab:main}, it is evident that models produced through different training methods exhibit varying degrees of robustness against general perturbations. 
The original CheckGPT model, which only trained the top-level LSTM with frozen RoBERTa, showed a limited capacity to withstand minor perturbations. Substituting some words with BERT yielded a substantial improvement, increasing $\Delta$Acc by 24\%. On the other hand, the HC3 with full parameter fine-tuning appeared more robust.
Notably, considering the relative trends, it becomes apparent that using a language model to generate candidate words and sentences demonstrates greater adversarial performance compared to heuristic replacement methods. 
Moreover, employing a prompt to guide LLMs in rewriting the text proves to be more versatile. Even without prompt engineering, it can still achieve notable adversarial effects on all detection models.

\begin{table*}[t]
\centering         
\resizebox{0.99\linewidth}{!} {%
\begin{tabular}{l|*{4}{c}|*{4}{c}|*{2}{c}}
\toprule
\multirow{2}{*}{\bfseries Model} & \multicolumn{4}{c}{\textbf{White-box Attack on HC3}} & \multicolumn{4}{c}{\textbf{Black-box Attack on CheckGPT}} & \multicolumn{2}{c}{\textbf{Text Quality}}\\
& {\bfseries AUC}~$\downarrow$ & {\bfseries PPV}~$\downarrow$ & {\bfseries TNR}~$\downarrow$ & {\bfseries $\bm{\Delta}$Acc}~$\uparrow$ & {\bfseries AUC}~$\downarrow$ & {\bfseries PPV}~$\downarrow$ & {\bfseries TNR}~$\downarrow$ & {\bfseries $\bm{\Delta}$Acc}~$\uparrow$ & {\bfseries $\bm{\Delta}$flesch\%} & {\bfseries $\bm{\Delta}$ppl} \\
\midrule
{HMGC} & 51.06 & 68.29 & 2.70 & 97.29 & 76.64 & 68.35 & 53.57 & 46.35 & -7.74 & 6.17 \\
{HMGC.$_{-USE}$} & 49.96 & 67.80 & 0.50 & 99.50 & 75.99 & 67.76 & 52.29 & 47.64 & -7.76 & 6.36 \\
{HMGC.$_{-MPR}$} & 50.82 & 68.18 & 2.21 & 97.79 & 76.65 & 68.36 & 53.59 & 46.33 & -7.75 & 6.17 \\
{HMGC.$_{-POS}$} & 50.04 & 67.84 & 0.65 & 99.35 & 77.95 & 69.59 & 56.20 & 43.72 & -6.36 & 5.86 \\
{HMGC.$_{-PPL}$} & 50.46 & 68.02 & 1.49 & 98.51 & 80.64 & 72.29 & 61.57 & 38.34 & -6.08 & 5.66 \\
\bottomrule
\end{tabular}
}
\caption{Attack performance and text quality comparisons between HMGC and its ablations.}
\label{tab:ablation}
\end{table*}

\begin{figure}[t]
\centering
\includegraphics[scale=0.3]{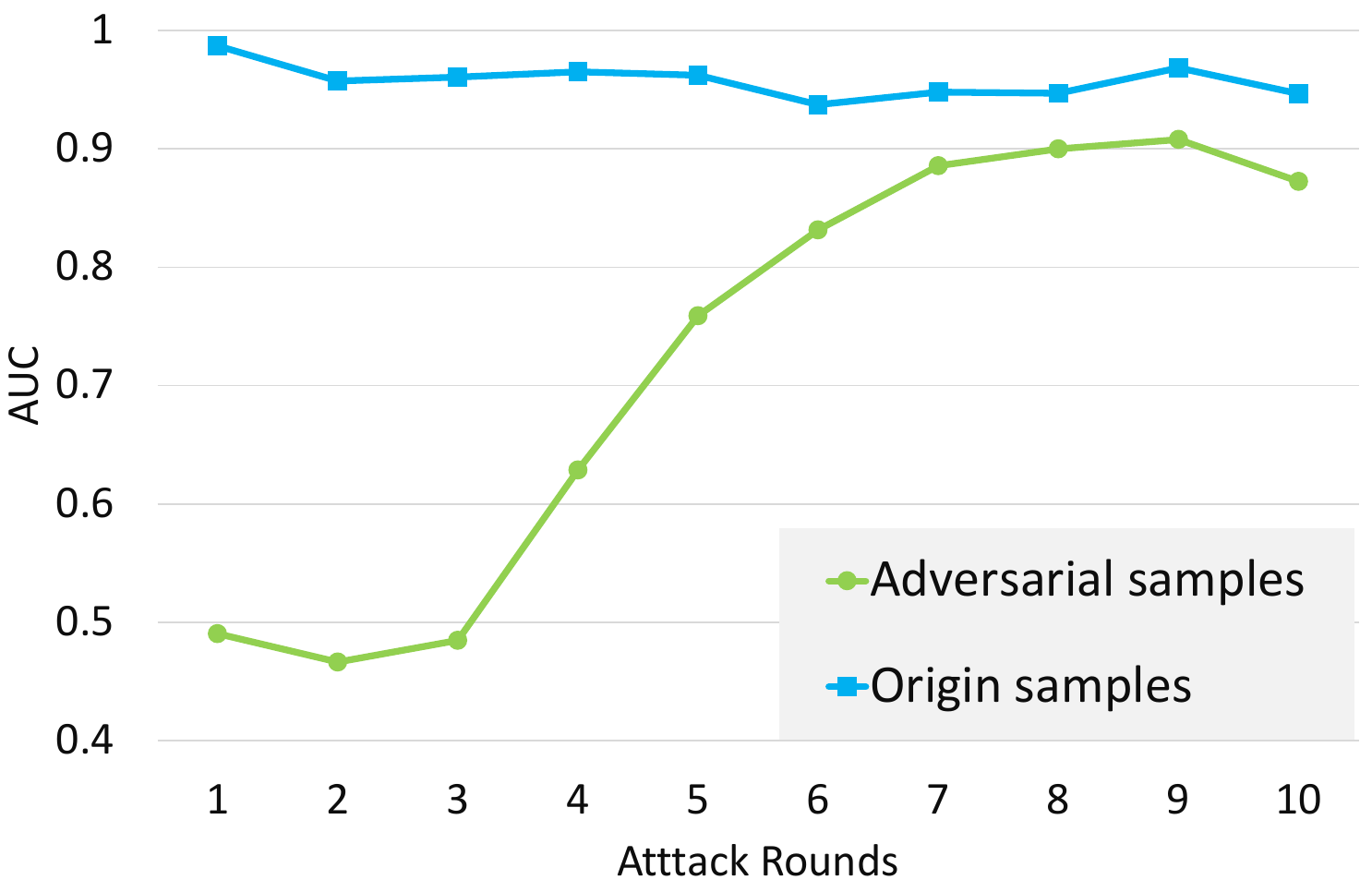}
\caption{Detection performance across different attack rounds. As the attack rounds intensify, the detector builds resilience against this form of attack.} 
\label{fig:dynamic} 
\end{figure}

\sparagraph{Adversarial learning can effectively enhance detector performance.} 
As depicted in Table~\ref{tab:dynamic}, the increase in the number of rounds of dynamic adversarial learning, which represents the continued training for the detector with adversarial samples, positively correlates with the detector's robustness. Meanwhile, this enhancement comes at the cost of increased time required to perform an adversarial attack.
Specifically, in the initial 3 rounds of attacks, the detector remains relatively vulnerable, with its AUC index dropping to approximately 0.5 whereas the time required for each attack approximately doubles. 
Subsequently, equilibrium is reached after roughly 7 rounds of attacks, corresponding to iterative learning from 50,000 adversarial examples. At this point, the attacker can produce an attack every 30 seconds, albeit with only about a 10\% success rate. This trend is visually illustrated in Figure~\ref{fig:dynamic}. 
Despite the improved robustness, the detector remains impractical for real-world applications, as it still yields an error rate of over 10\% when classifying adversarial content as human-written.
In summary, these experimental results demonstrate the positive impact of iterative adversarial learning in dynamic scenarios on enhancing detector robustness. 
Since the current detector does not incorporate adversarial attack considerations in its model design, our research contributes valuable insights for the development of future detection models.

\sparagraph{Trade-off: evasion of detection or preservation of original semantics.}
We conducted an ablation analysis to examine the key modules in the HMGC model design. From the results in Table~\ref{tab:ablation}, we observe the following: 
1) In the white-box attack setting, which is relatively straightforward, the ablation analysis of the model causes only minor fluctuations in attack accuracy, typically within 1 or 2 percentage points. These variations may stem from the random factors involved in the attack experiments.
2) The black-box attack setting effectively demonstrates the significance of the perplexity word importance we proposed. When the module is removed as HMGC.$_{-PPL}$, the attack success rate decreases by 8\%, indicating the effectiveness of word perplexity in the ADAT task.
3) Ablation results for other modules show that the attack success rate is directly proportional to the language perplexity of the adversarial text. For instance, when the USE constraint is removed as HMGC.$_{-USE}$, the attack success rate increases by approximately 1\%, but the corresponding language perplexity also rises by 0.2.
From these observations, we can deduce that an effective strategy to evade AI-text detection is introducing external noise to increase text perplexity. 
However, this approach may face the challenge of semantic shifts between the original text and the adversarial text. 
Balancing these factors should be a crucial consideration in future research on AI-text detection.

\section{Conclusion}
In this paper, we introduce the Adversarial Detection Attack on AI-Text (ADAT) task, which includes two attack settings: white-box and black-box attacks. Furthermore, we propose a novel approach involving adversarial learning in dynamic scenarios to enhance the resistance of detection models against adversarial attacks.
Our algorithm demonstrations and experimental results prove the vulnerability of the current design in detection models, revealing their susceptibility to even minor perturbations that can effectively disrupt the final prediction results. 
To perform effective adversarial attacks, we present the Humanizing Machine-Generated Content (HMGC) framework, which emulates the interactive attack process between an attacker and a detector, continuously refining the attack strategy based on the rewards provided by the detector.
Our proposed approach, supported by extensive experimental results, not only highlights the vulnerabilities in existing AI-text detection methods but also sheds light on the risks and directions for future research in the AI-detection domain.

In future work, we plan to expand our framework to support sentence-level and document-level substitutions to produce more fluent adversarial texts. 
Concurrently, we will refine the adversarial learning approach in dynamic scenarios to train a more robust, stable, and versatile AI-text detector.
Moreover, beyond the ADAT tasks, exploring more general content correction technology for the AI-text also appears to be a promising direction for further research.

\section*{Acknowledgements}
This work is supported by the National Natural Science Foundation of China (62272439), and the Fundamental Research Funds for the Central Universities.

\bibliography{main}

\appendix



\end{document}